\documentclass[times,review,10pt]{elsarticle}

\usepackage{booktabs}
\usepackage{multirow}
\usepackage{graphicx}
\usepackage{xcolor} 
\usepackage{url}
\usepackage{makecell}

\usepackage{amssymb}
\usepackage{amsmath}
\journal{Pattern Recognition}

\begin{document}

\begin{frontmatter}


\title{A Semi-supervised Physics-Aware Triple-Stream Underwater Image Enhancement Network} %% Article title

\author[1]{Shixuan Xu\fnref{fn1}}
\ead{xushixuan@stu.ouc.edu.cn}

\author[1]{Hao Qi\fnref{fn1}}
\ead{qihao@stu.ouc.edu.cn}

\author[2]{Wei Wang}
\ead{wangwei29@mail.sysu.edu.cn}

\author[2]{Chao Huang}
\ead{huangch253@mail.sysu.edu.cn}

\author[3]{Jie Wen}
\ead{jiewen\_pr@126.com}

\author[1]{Junyu Dong}
\ead{dongjunyu@ouc.edu.cn}

\author[1]{Xinghui Dong\corref{cor1}}
\ead{xinghui.dong@ouc.edu.cn}

\cortext[cor1]{Corresponding author}

\affiliation[1]{organization={Ocean University of China},
            addressline={State Key Laboratory of Physical Oceanography, Faculty of Information Science and Engineering},
            city={Qingdao},
            postcode={266100},
            state={Shandong},
            country={China}}

\affiliation[2]{organization={Shenzhen Campus of Sun Yat-sen University},
            addressline={School of Cyber Science and Technology},
            city={Shenzhen},
            state={Guangdong},
            country={China}}

\affiliation[3]{organization={Harbin Institute of Technology, Shenzhen},
            addressline={Shenzhen Key Laboratory of Visual Object Detection and Recognition},
            city={Shenzhen},
            state={Guangdong},
            country={China}}

%% Funding & equal contribution
\fntext[fn1]{These authors contributed equally.}

\tnotetext[Github]{Code and models are available at https://github.com/INDTLab/PATS-UIENet.}

%% Abstract
\begin{abstract}
Underwater images normally suffer from degradation due to the transmission medium of water bodies. Both traditional prior-based approaches and deep learning-based methods have been used to address this problem. However, the inflexible assumption of the former often impairs their effectiveness in handling diverse underwater scenes, while the generalization of the latter to unseen images is usually weakened by insufficient data. In this study, we leverage both the physics-based Image Formation Model (IFM) and deep learning techniques for Underwater Image Enhancement (UIE). To this end, we propose a novel Physics-Aware Triple-Stream Underwater Image Enhancement Network, i.e., PATS-UIENet, which comprises a Direct Signal Transmission Estimation Stream (D-Stream), a Backscatter Signal Transmission Estimation Stream (B-Stream) and an Ambient Light Estimation Stream (A-Stream). This network performs UIE by explicitly predicting three degradation variables defined by a revised IFM. We also adopt an IFM-inspired semi-supervised learning framework, which exploits both the labeled and unlabeled images, to address the issue of insufficient data. To our knowledge, this study makes the first effort to jointly apply the physics-aware deep network and the IFM-inspired semi-supervised learning technique to the UIE task. Our method performs better than, or at least comparably to, sixteen baselines across six testing sets in the degradation estimation and UIE tasks. These promising results should be due to the fact that the proposed method can not only model the degradation but also learn the characteristics of diverse underwater scenes.
\end{abstract}


\begin{highlights}
\item Designing a triple-stream architecture, including direct transmission, backscatter transmission, and ambient light estimation streams, for physics-guided underwater degradation modeling. 
\item Proposing a Physics-Aware Triple-Stream Underwater Image Enhancement Network (PATS-UIENet), explicitly estimating degradation variables of the Image Formation Model (IFM) through three dedicated streams. 
\item Developing an IFM-inspired semi-supervised learning framework, effectively leveraging both labeled and unlabeled data to improve generalization under limited supervision.
\end{highlights}

%% Keywords

\end{frontmatter}



\section{Introduction}
\label{intro}
\begin{figure}[t]
\centering
\includegraphics[width=0.99\linewidth]{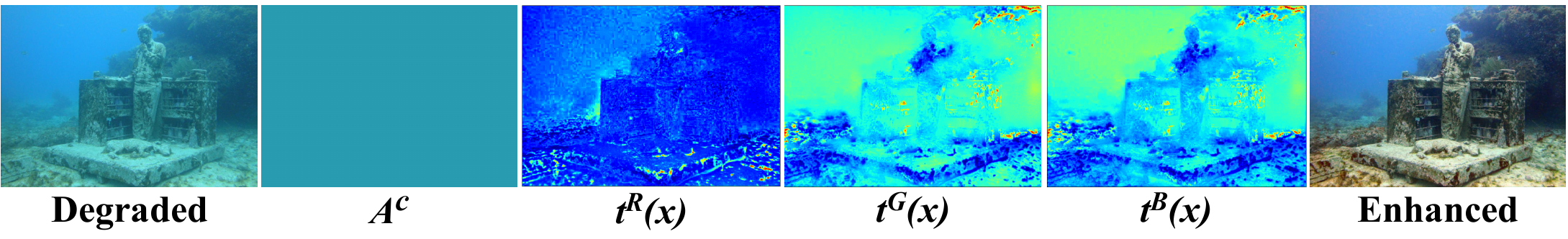}
\caption{Examples of a degraded underwater image, the corresponding ambient light image $A^c$ and three transmission maps $t^c(x)$ ($c \in \{\mathrm{R},\mathrm{G},\mathrm{B}\}$), and the enhanced image produced using the predicted variables of the IFM \cite{IFM}.}
\label{fig:ifm_show}
\end{figure}

The images captured in the underwater environment play important roles in ocean exploration \cite{UWCNN, pr3}. However, these images normally suffer from different degradation, due to the wavelength-dependent light absorption and light scattering caused by the underwater transmission medium. According to the Image Formation Model (IFM) \cite{IFM}, the process of image degradation can be formulated as follows:
\begin{equation}
\label{euq:ifm}
I^c(x) = J^c(x)t^c(x) + (1-t^c(x))A^c,
\end{equation}
where $I^c(x)$ ($c\in\{\mathrm{R},\mathrm{G},\mathrm{B}\}$) is a degraded image, $J^c(x)$ is the underlying clean image, $t^c(x) = \exp(-\beta^c d(x))$ denotes the transmission maps, $\beta^c$ is the attenuation factor for each channel, $d(x)$ is the scene distance map and $A^c$ is the ambient light. (See Fig. \ref{fig:ifm_show} for examples). In this context, $J^c(x)t^c(x)$ describes the distance-dependent color distortion, while $(1-t^c(x))A^c$ represents the scattering of the ambient light which decreases the contrast and lessens the visibility of the image. Despite the IFM had been widely used, its flaws in accurately modeling the degradation process was highlighted \cite{revised_IFM}.

Considering that the degradation may interfere with downstream tasks, the design of effective UIE methods is critical yet challenging. Traditional prior-based methods \cite{DCP, UDCP, GDCP, WCID, hl, water_hl, retinex, mmle, hlrp, fusion1, hs2cm2a} typically rely on a certain form of \textit{prior} information. Some prior-based methods \cite{DCP, UDCP, GDCP} aim at estimating the parameters of the IFM, but usually struggle with adapting to diverse underwater scenes or estimating valid parameters when the scene violates the prior assumption \cite{GDCP}. Other prior-based methods \cite{retinex, mmle, hlrp, hs2cm2a} directly enhance the quality of underwater images without explicitly modeling the degradation process, yet their limited adaptability often leads to artifacts in complex scenes.

In contrast, deep learning-based UIE approaches \cite{ucycle_gan, fast_gan, ushape, waternet, ucolor, usuir, puie, uranker, ccmsrnet, semi-uir, ddformer, uwformer} directly learn from degraded images and thus alleviate the limitations of static priors. Recent studies have further incorporated physical models \cite{pugan, phish, bluenet} or multi-modal and frequency-domain guidance \cite{watercyclediffusion, ws-uienet} into learnable enhancement frameworks. However, constructing large labeled underwater data sets is time-consuming and expensive. Unsupervised learning methods \cite{homology} reduce the need for paired data but often suffer from slow inference and color distortion, while recent semi-supervised methods \cite{semi-uir, ddformer, uwformer} leverage both labeled and unlabeled data to improve performance but still face challenges when training data is insufficient or noisy. 

Despite the progress has been made by the existing methods, two challenges remain. First, underwater degradation is governed by spatially varying and coupled physical factors, making the accurate estimation of physically meaningful variables difficult. Second, a supervised or unsupervised approach normally struggles with the insufficient or noisy data due to the complexity of real underwater degradation, leading to overfitting or weak generalization.

To address these challenges, we propose a Physics-Aware Triple-Stream Underwater Image Enhancement Network, i.e., PATS-UIENet, which explicitly incorporates the revised IFM \cite{revised_IFM} into the deep neural network that we deliberately design. Unlike existing physics-based methods \cite{DCP, UDCP, GDCP}, our network explicitly predicts the degradation variables defined by a physical model using three streams, including a Direct Signal Transmission Estimation Stream (D-Stream), a Backscatter Signal Transmission Estimation Stream (B-Stream) and an Ambient Estimation Stream (A-Stream). Due to the modeling of the physical degradation process, our network is superior to existing UIE methods by exploiting the strengths of both the theoretical interpretability of physics-based approaches and the powerful feature representation ability of deep neural networks.

We further introduce an IFM-inspired semi-supervised learning framework, including a bi-directional supervised scheme and an unsupervised scheme, for the purpose of overcoming the limitation of insufficient data. The supervised scheme effectively utilizes the limited labeled data by providing explicit degradation guidance, while the unsupervised scheme exploits the abundant unlabeled data to improve generalization. Therefore, our framework can be better trained than the existing supervised approaches \cite{UWCNN, ucycle_gan, waternet, ucolor, usuir, puie, uranker, ccmsrnet}, unsupervised methods \cite{homology} and semi-supervised approaches \cite{semi-uir, ddformer, uwformer} using the same amount of data. As a result, the generalization of the model trained using our framework is stronger than that trained using existing supervised, unsupervised and semi-supervised approaches. 

To our knowledge, this study makes the first effort to jointly apply the physics-aware deep network and the IFM-inspired semi-supervised learning technique to the UIE task. Our main contributions can be summarized as threefold. 
\begin{itemize}
\item{We introduce PATS-UIENet, which uses three dedicated streams to predict the direct transmission, backscatter transmission, and ambient light under the guidance of the revised IFM.}
\item{We propose an IFM-inspired semi-supervised learning framework, which addresses the issues of data insufficiency and training instability by exploiting the merits of both the supervised and unsupervised learning methods.}
\item{We conduct a series of comparative experiments on six underwater testing sets along with sixteen baselines. The results not only validate the effectiveness of our method but also provide benchmarks for future research.}
\end{itemize}

The remainder of this paper is organized as follows. The related literature is reviewed in Section \ref{realted_work}. We introduce the proposed method in Section \ref{methods}. The experimental settings and results are reported in Sections \ref{experiment} and \ref{results} respectively. Finally, our conclusion is drawn in Section \ref{conclusion}.

\section{Related Work}
\label{realted_work}
\subsection{Revised Image Formation Model}
As pointed out by Akkaynak and Treibitz \cite{revised_IFM}, the original IFM \cite{IFM} ignored the dependencies of the backscatter coefficient on the ambient light and the optical property of water bodies. They assumed that the attenuation coefficients of the direct signal and the backscattered signal are the same. In this case, the original IFM \cite{IFM} cannot adequately describe the degradation process of underwater images. It has been demonstrated that the attenuation factors of the direct signal and the backscatter signal are different while the attenuation factor of the backscatter signal is affected more severely by the ambient light \cite{revised_IFM}.

Akkaynak and Treibitz \cite{revised_IFM} further proposed a revised IFM, which can be expressed as:
\begin{equation}
\label{euq:revised_ifm}
I^c(x) = J^c(x)\exp\left(-\beta_c^{\mathrm{D}}(\mathbf{D}) \cdot z\right)
+ \left(1 - \exp\left(-\beta_c^{\mathrm{B}}(\mathbf{B}) \cdot z\right)\right)A^c, 
\end{equation}
where $\textbf{D}=\{ z, \rho, E, S_c, \beta \}$ and $\textbf{B}=\{ E, S_c, b, \beta \}$ are two sets of parameters which affect the attenuation factors of the direct signal $\beta_c^D$ and the backscatter signal $\beta_c^B$, respectively, $z$ represents the distance between the scene and the camera, $\rho$ is the reflectance spectrum of the object, $E$ is the ambient light at a certain distance, $S_c$ is the camera response function, and $b$ and $\beta$ are the beam scattering and attenuation coefficients, respectively. 

In contrast to the original IFM \cite{IFM}, the revised IFM \cite{revised_IFM} offers the more comprehensive and physically accurate representation of underwater image degradation. Therefore, we design the proposed method on top of the revised IFM. For more details, please refer to the original publication \cite{revised_IFM}.

\subsection{Prior-Based Methods}

To recover clear underwater images, many methods \cite{DCP, UDCP, GDCP, WCID, hl, water_hl} used the \textit{prior} information to estimate the degradation parameters of the IFM. Although the Dark Channel Prior (DCP) \cite{DCP} was initially proposed for image dehazing, some researchers \cite{WCID} employed it for underwater image enhancement, due to the similarity between the degradation processes resulted from the foggy weather and underwater environment. However, the original DCP usually yielded erroneous estimations. Therefore, more studies were performed to improve it, such as Underwater Dark Channel Prior (UDCP) \cite{UDCP} and Generalization of the Dark Channel Prior (GDCP) \cite{GDCP}. Moreover, the Haze Line Prior (HL) \cite{hl} was utilized in some studies \cite{water_hl, water_hl_2}. Despite these methods were designed on top of the IFM, they normally made a rigid assumption about the underwater environment, which restricted the application of them to a specific underwater scenario.

On the other hand, some approaches used more general priors \cite{RGHS, retinex, mmle, hlrp, hs2cm2a} to enhance the quality of underwater images without taking the IFM into account, including Retinex-based methods \cite{retinex}, image fusion techniques \cite{fusion1}, and histogram similarity-oriented color compensation with multiple attribute adjustment \cite{hs2cm2a}. These approaches usually enhanced the contrast and produced more color-balanced results. However, they probably struggle with processing globally inhomogeneous degraded images and may introduce artifacts, such as halos and color casts, due to the lack of the knowledge of underwater scenes.

\subsection{Learning-Based Methods}

Thanks to the powerful representation learning ability and large-scale training data, deep learning-based methods greatly boosted the development of computer vision in both high-level and low-level vision tasks. However, the application of deep learning to underwater image enhancement is much less than other tasks. This dilemma should be attributed to the difficulty in collecting a large number of underwater images and the corresponding clean counterparts. Recent studies have incorporated physical models \cite{pugan, phish, bluenet} or multi-modal and frequency-domain guidance \cite{watercyclediffusion, ws-uienet} into learnable UIE frameworks. Visual-textual interaction has also been investigated in remote sensing visual question answering through VMGN \cite{vmgn}, demonstrating the potential of multimodal guidance for imaging tasks.

In contrast to the lack of labeled underwater images, it is easier to collect unlabeled underwater images. Existing studies \cite{ucycle_gan} sourced real-world underwater images and categorized them according to the degree of degradation. Unsupervised learning methods \cite{usuir} were developed on top of these data by learning the mapping from severely degraded images to slightly degraded images. However, these approaches were difficult to train and tended to produce unstable results. Recently, semi-supervised learning methods \cite{semi-uir, ddformer, uwformer} have been developed, using both labeled and unlabeled underwater images. Although promising results had been derived, these methods mainly focused on exploiting unlabeled data without explicitly incorporating the underwater image formation process into both network design and training.

The above-mentioned studies either are not robust to diverse underwater scenes, or lack the knowledge of underwater images. We are hence motivated to exploit both the IFM-inspired semi-supervised learning technique and the physics-aware deep neural network, to address these issues.

\section{Methodology}
\label{methods}

Considering the importance of physical principles to the UIE task, we propose a Physics-Aware Triple-Stream Underwater Image Enhancement Network (PATS-UIENet). This network contains a Direct Signal Transmission Estimation Stream (D-Stream), a Backscatter Signal Transmission Estimation Stream (B-Stream) and an Ambient Light Estimation Stream (A-Stream), which predict the three degradation variables of the revised IFM \cite{revised_IFM}, respectively. To address the challenge of the lack of labeled real-world underwater images, we further adopt an IFM-inspired semi-supervised learning framework, which consists of a bi-directional supervised scheme and an unsupervised scheme. Compared to the supervised or unsupervised method, the PATS-UIENet can be better trained using this framework with both the labeled and unlabeled real-world images while the generalization of the model trained is thus stronger, due to the complementary action of the two schemes.

\subsection{Physics-Inspired Design}

Motivated by the revised IFM \cite{revised_IFM}, we propose a physics-aware underwater image enhancement network on top of this model. The three streams of this network correspond to the direct signal transmission map, backscatter signal transmission map, and ambient-light components. To reduce the complexity of the network, we simplify Eq. (\ref{euq:revised_ifm}) as follows:
\begin{equation}
\label{euq:revised_ifm_2}
I^c(x) = J^c(x)t_{\mathrm{D}}^c(x)
+ \left(1 - t_{\mathrm{B}}^c(x)\right)A^c,
\end{equation}
where $I^c(x)$ ($c\in\{\mathrm{R},\mathrm{G},\mathrm{B}\}$) is the degraded image, $J^c(x)$ is the underlying clean image, $t_{\mathrm{D}}^c(x)$ and $t_{\mathrm{B}}^c(x)$ represent the direct and backscatter signal transmission maps, respectively, and $A^c$ denotes the ambient light. To predict these variables, including $t_{\mathrm{D}}^c(x)$, $t_{\mathrm{B}}^c(x)$ and $A^c$, we design a D-Stream, a B-Stream and an A-Stream, respectively. 

We also introduce an IFM-inspired semi-supervised learning framework by leveraging the explicitly predicted variables of the revised IFM \cite{revised_IFM}. This framework consists of a bi-directional supervised learning scheme and an unsupervised learning scheme, which addresses the challenge of rare labeled real-world underwater images. Compared to the supervised or unsupervised methods, our PATS-UIENet can take advantage of both labeled and unlabeled real-world images for the more effective training operation. Furthermore, the complementary action between the two schemes enhances the generalization ability of the model trained.

\subsection{PATS-UIENet}

\begin{figure}[t]
\centering    
\includegraphics[width=0.99\textwidth]{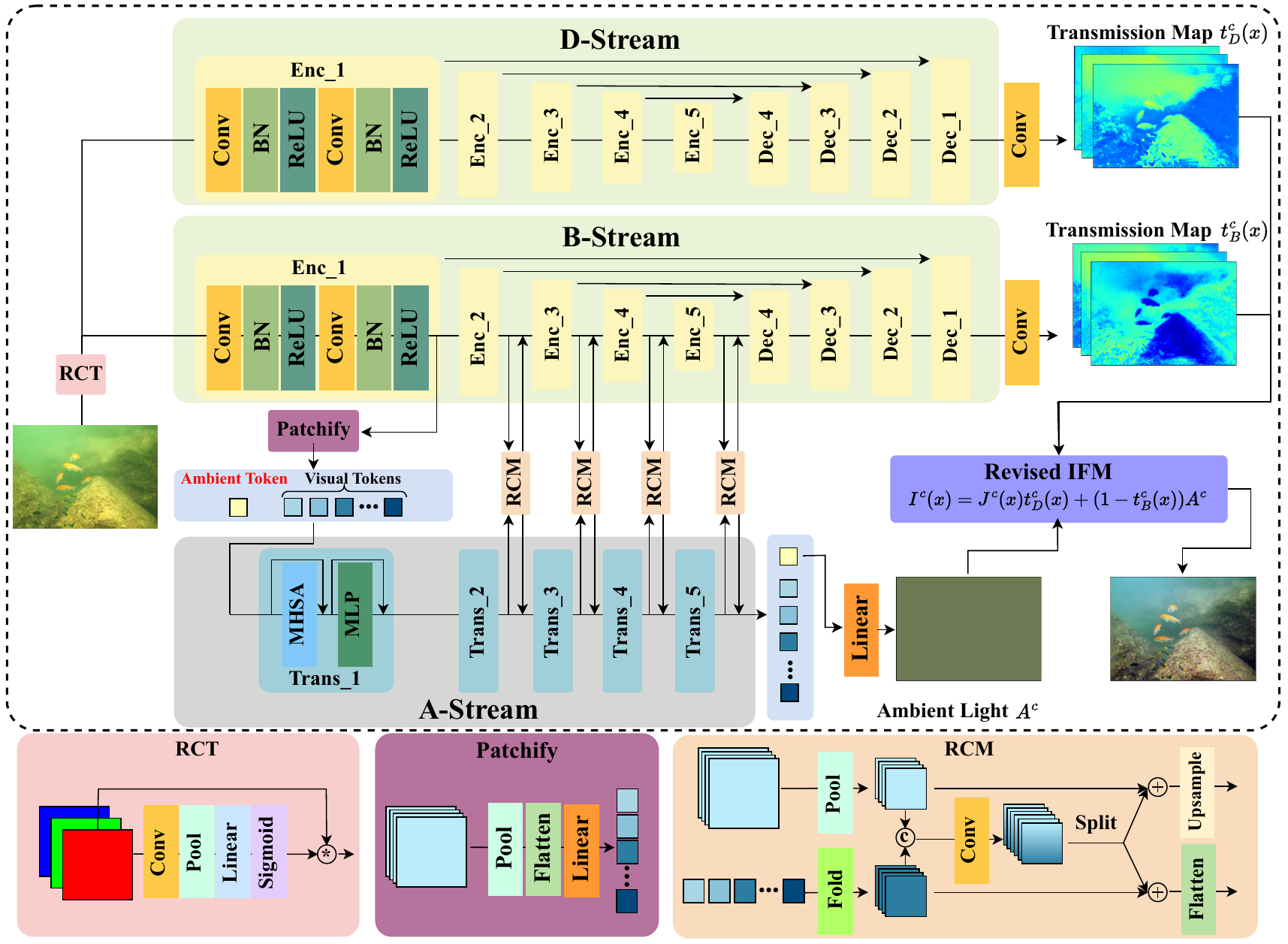}
\caption{The architecture of the proposed PATS-UIENet, which comprises three individual streams, namely, D-Stream, B-Stream and A-Stream, to estimate the degradation variable of the revised IFM \cite{revised_IFM}.}
\label{fig:arch}
% \vspace{-10pt}
\end{figure}

As illustrated in Fig. \ref{fig:arch}, the PATS-UIENet contains two encoder-decoder style streams with the same structure, i.e., D-stream and B-stream, and a Transformer-based A-stream. The three streams are used to estimate the direct signal transmission map, the backscatter signal transmission map and the ambient light, respectively. According to Eq. (\ref{euq:revised_ifm_2}), both the direct signal transmission $t_D^c(x)$ and the backscatter signal transmission $t_B^c(x)$ are spatially varying pixel-wise maps. Estimation of them requires the preservation of local structures and multi-scale spatial information. As a result, the D-Stream and B-Stream are constructed using CNN-based encoder-decoder networks with skip connections, which are suitable for dense pixel-wise prediction.

In contrast, the ambient light $A^c$ in Eq. (\ref{euq:revised_ifm_2}) is modeled as a scene-level variable and does not explicitly vary with the spatial coordinate $x$. Estimation of it hence relies more on the global information of the input image. Thus, a Transformer-based architecture is adopted for the A-Stream due to its ability to model long-range dependencies and global characteristics. Besides, the backscatter component is jointly determined by the backscatter signal transmission and the ambient light, i.e., $\left(1-t_B^c(x)\right)A^c$. Therefore, the B-Stream also exchanges features with the A-Stream through the RCMs to incorporate global ambient light information even though it retains a CNN-based architecture to estimate a spatially varying transmission map.

\textbf{Red Channel Tuner (RCT).}  Inspired by the channel attention mechanism \cite{se_net}, we propose a compact RCT module in order to adaptively emphasize the information contained in the red channel. This module first uses a convolutional layer to extract basic features from the input image, and then use the Global Average Pooling (GAP) to obtain a global representation. A fully-connected layer and the \textit{Sigmoid} activation function are further used to transform this representation to a tuning weight. The weight is finally used to scale the red channel of the image.

\textbf{D-Stream.} D-stream is designed to estimate the direct signal transmission map of an underwater image. According to Eq. (\ref{euq:revised_ifm_2}), the transmission map varies at the locations of different pixels. In this case, the network should be able to generate fine-grained representations. Hence, we build the D-Stream using a CNN-based encoder-decoder network, due to its strong ability to learn local characteristics, which helps preserve edges and fine details. The encoder comprises five consecutive blocks, denoted as $Enc\_1$ to $Enc\_5$. As a result, multi-scale feature maps are derived using the encoder. Given $Enc\_5$ is used as the bottleneck, the decoder contains four blocks, denoted as $Dec\_1$ to $Dec\_4$, symmetrically. Skip connection is used to pass the feature maps produced by an encoder block to the corresponding decoder block at the same level, which is useful for restoring the fine-grained spatial structure. The use of multi-scale features and skip connections enhances the ability of the network to restore the spatial structure of the scene. The output of the last decoder block is fed into a convolutional layer. Three transmission maps are produced in terms of different color channels. In essence, D-Stream is specifically focused on modeling the directly transmitted light, which normally contains the spatial structure of the scene.

\textbf{B-Stream.} B-stream is responsible for modeling the degradation caused by the backscattering in underwater environments. Although it shares the same CNN-based architecture as D-Stream for training stability, B-stream differs in functionality and processing strategy. Unlike direct signals, backscatter signals exhibit strong global interference and require a context-aware estimation. To this end, we pass the output of the first encoder block $Enc\_1$ through a \textit{Patchify} processing and feed it into the A-Stream, leveraging the global modeling capability of Transformer to enhance the estimation accuracy. Moreover, B-Stream is connected with the A-Stream via Residual Communication Modules (RCMs), enabling bidirectional feature exchange. Compared to the D-Stream, B-Stream is focused on modeling the global degradation rather than recovering the local scene content.

\textbf{A-Stream.} Ambient light reflects the overall luminance in the underwater scene and is generally spatially homogeneous. To effectively model this global characteristic, we employ a Transformer-based design instead of CNNs, because Transformer is well-suited for capturing long-range dependencies and global characteristics through the self-attention mechanism. A-Stream consists of five Transformer blocks, denoted as $Trans\_1$ to $Trans\_5$. Specifically, the output of $Enc_1$ in the B-Stream is processed by \textit{Patchify} to generate a sequence of visual tokens, and a learnable ``Ambient" token is prepended to the sequence. Through self-attention, the Ambient token interacts with all visual tokens and aggregates global information related to ambient light. In addition, the RCMs enable bidirectional feature exchange between the A-Stream and the multi-scale features of the B-Stream. After the last Transformer block, the token sequence is normalized, and the updated Ambient token at the first position is fed into a fully-connected layer followed by a sigmoid activation function to estimate the ambient light $\hat A^c$ for the three color channels. The estimated ambient light is then expanded to all spatial locations and combined with the backscatter signal transmission map according to Eq. (\ref{euq:revised_ifm_2}). It should be noted that the Ambient token is a dedicated latent representation for ambient light estimation rather than the ambient light value.

\textbf{Residual Communication Module (RCM).} According to the revised IFM \cite{revised_IFM}, the degradation can be considered as the outcome caused by both the transmission and the ambient light. In this situation, the backscatter signal transmission estimation stream and the ambient light estimation stream will learn some common features. %from the degraded images. 
Therefore, the feature exchange between both the streams is likely to boost the training of each stream. Recently, the complementary action of CNNs and Transformers has been shown \cite{conformer}. We are inspired to propose the RCM for the sake of bridging the two streams. Specifically, the tokens produced by a block in the A-stream are folded into a set of feature maps while the feature maps generated by a block of the encoder of the B-stream are downsampled to the shape of these maps. Then both sets of maps are concatenated along the channel dimension and are fed into a convolutional layer. Finally, the resultant maps are split into two sets along the channel dimension. Each set is added with the original feature maps of the related stream.

\subsection{IFM-Inspired Semi-supervised Learning Framework}

\begin{figure*}[t]
\centering
\includegraphics[width=0.99\textwidth]{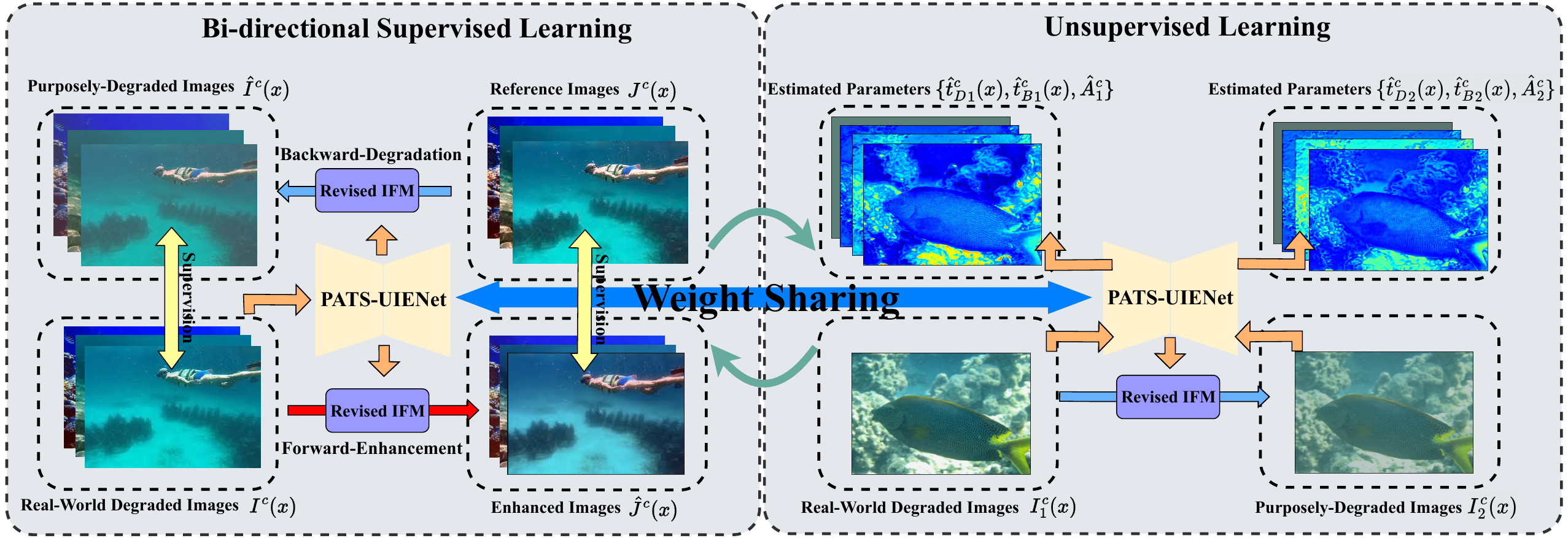}
\caption{The proposed semi-supervised learning framework, which comprises a bi-directional supervised learning scheme and an unsupervised learning scheme, used for training our PATS-UIENet.}
\label{fig:semi-supervised}
% \vspace{-10pt}
\end{figure*}

To overcome the challenge of insufficient real-world underwater images, we adopt a semi-supervised learning framework (see Fig. \ref{fig:semi-supervised}), inspired by the revised IFM \cite{revised_IFM} theory, embedding physical constraints directly into the learning process to effectively leverage unlabeled data. This framework includes a bi-directional supervised scheme and an unsupervised scheme. The bi-directional scheme contains a forward-enhancement module, which learns from reference images and uses the predicted variables of the revised IFM to perform UIE, and a backward-degradation module, which uses the predicted variables to degrade the reference images to their real-world degraded counterparts. As a result, the bi-directional supervised scheme can learn more reliable degradation representations using limited real-world data than that learned using a supervised method with a large number of synthetic data. Also, the generalization of the model trained is stronger than that trained using a supervised method with limited real-world data. On the other hand, the unsupervised scheme exploits the unlabeled real-world underwater images that we collect. A second set of images with different degrees of degradation are generated from these images based on the revised IFM \cite{revised_IFM}. Two sets of variables can be estimated using both sets of images, respectively. Each set is utilized as the reference for the other set because their IFMs are related.

\textbf{Bi-directional Supervised Learning}. Given a set of labeled training data, we first use the predicted variables to obtain enhanced results $\hat J^c(x)$. Then we perform the forward-enhancement learning module by minimizing the following loss function:
\begin{equation}
\label{euq:forward}
\mathcal{L}_{\mathrm{fwd}} =
\left\lVert J^{c}(x) - \hat{J}^{c}(x) \right\rVert_2^2.
\end{equation}
Due to the ill-posed nature of the revised IFM, however, only $\mathcal{L}_{\mathrm{fwd}}$ may be insufficient for ensuring the validity of the predicted variables. Therefore, we propose a backward-degradation learning module which degrades clear reference images using the predicted variables to their degraded counterparts $\hat I^c(x)$. The supervision over the backward-degradation is achieved by minimizing the loss function:
\begin{equation}
\label{euq:backward}
\mathcal{L}_{\mathrm{bwd}} =
\left\lVert I^{c}(x) - \hat{I}^{c}(x) \right\rVert_2^2.
\end{equation}

Inspired by the Retinex \cite{retinex} theory, we further apply Gaussian blur to input images in order to derive the hint of the global ambient light, which is able to boost the training of the A-stream. Given the ambient light $\hat A^c$ estimated using the A-stream, we minimize the following loss function:
\begin{equation}
\label{euq:aloss_bisup}
\mathcal{L}_{\mathrm{A}\text{-}\mathrm{sup}} =
\left\lVert I^{c}(x)\ast \mathbf{G} - \hat{A}^{c} \right\rVert_2^2,
\end{equation}
where $\mathbf{G}$ stands for a Gaussian kernel and $\ast$ is the convolution operation. Finally, the bi-directional supervised learning scheme is conducted as: 
\begin{equation}
\label{euq:bi-supervision}
\mathcal{L}_{\mathrm{sup}} =
\mathcal{L}_{\mathrm{fwd}}
+ \lambda_1 \mathcal{L}_{\mathrm{bwd}}
+ \lambda_2 \mathcal{L}_{\mathrm{A}\text{-}\mathrm{sup}},
\end{equation}
where $\lambda_1$ and $\lambda_2$ are used to balance different loss functions. Compared with the supervised methods \cite{waternet, ucolor}, this scheme exploits the limited labeled underwater images better. 

\textbf{Unsupervised Learning.} To leverage unlabeled underwater images, we adopt an unsupervised learning scheme based on the physical modeling capability of the revised IFM \cite{revised_IFM} rather than using generative networks \cite{ucycle_gan}, which are difficult to train and often produce unstable results. This scheme adaptively constructs pairs of images with different levels of degradation from the unlabeled data.

Given an unlabeled real-world degraded image $I^{c}_1(x)$, the revised IFM can be expressed as:
\begin{equation}
\label{euq:degrade1}
I^{c}_{1}(x) = J^{c}_{1}(x)t_{\mathrm{D}_{1}}^{c}(x) + (1-t_{\mathrm{B}_{1}}^{c}(x))A^{c}_{1}.
\end{equation}
A new degraded image $I^{c}_{2}(x)$ can be purposely derived according to:
\begin{equation}
\label{euq:degrade2}
I^{c}_{2}(x) = \alpha I^{c}_{1}(x) + (1-\alpha)(1-t_{\mathrm{B}_{1}}^{c}(x))A^{c}_{1},
\end{equation}
where $\alpha \in (0,1)$ is a controlled factor which decides the degradation extent. Substituting $I^{c}_{1}(x)$ in Eq. (\ref{euq:degrade1}) into Eq. (\ref{euq:degrade2}), $I^{c}_{2}(x)$ can be then expressed as:
\begin{equation}
\label{euq:degrade3}
I^{c}_{2}(x) = J^{c}_{1}(x)(\alpha t_{\mathrm{D}_{1}}^{c}(x))+ (1-t_{\mathrm{B}_{1}}^{c}(x))A^{c}_{1}.
\end{equation}
Eq. (\ref{euq:degrade3}) also satisfies the revised IFM but presents a more severe degradation than the $I^{c}_{1}(x)$, generated from the same underlying clear image $J^{c}_{1}(x)$, because $\alpha t_{\mathrm{D}_{1}}^{c}(x) < t_{\mathrm{D}_{1}}^{c}(x)$ always holds true. Therefore, the transmission maps $\hat t_{\mathrm{D}_{2}}^{c}(x)$ and $\hat t_{\mathrm{B}_{2}}^{c}(x)$ estimated using the PATS-UIENet from image $I^{c}_{2}(x)$ will be close to $\alpha \hat t_{\mathrm{D}_{1}}^{c}(x)$ and $\hat t_{\mathrm{B}_{1}}^{c}(x)$, respectively, where $\hat t_{\mathrm{D}_{1}}^{c}(x)$ and $\hat t_{\mathrm{B}_{1}}^{c}(x)$ are the transmission maps predicted from $I^{c}_{1}(x)$. Correspondingly, two loss functions are defined as:
\begin{equation}
	\begin{aligned}
		& \mathcal{L}_{\mathrm{D}} = ||\hat t_{\mathrm{D}_{2}}^{c}(x) - \alpha \hat t_{\mathrm{D}_{1}}^{c}(x)||_2^2,		
	\end{aligned}
\end{equation}
\begin{equation}
	\begin{aligned}
		& \mathcal{L}_{\mathrm{B}} = ||\hat t_{\mathrm{B}_{1}}^{c}(x) - \hat t_{\mathrm{B}_{2}}^{c}(x)||_2^2.		
	\end{aligned}
\end{equation}
Similar to Eq. (\ref{euq:aloss_bisup}), the loss function for the ambient light $\hat A^c$ estimated by the A-stream is defined as:
\begin{equation}
\label{euq:aloss_unsup}
	\begin{aligned}		
\mathcal{L}_{\mathrm{A}\text{-}\mathrm{unsup}} =
\left\lVert I^{c}(x)\ast \mathbf{G} - \hat{A}^{c} \right\rVert_2^2.
	\end{aligned}
\end{equation}

Due to the lack of reference images, we use a non-reference loss function \cite{homology} based on the Gray-world prior \cite{gray_world} to constrain the enhancement of image $\hat J^c_1(x)$, which is expressed as:
\begin{equation}
\mathcal{L}_{\mathrm{gw}} =
\sum_{c \in \{\mathrm{R},\mathrm{G},\mathrm{B}\}}
\left\lVert
\operatorname{Avg}\left(J_1^c(x)\right) - 0.5
\right\rVert_2^2,
\end{equation}
where $\operatorname{Avg}(J_1^c(x))$ denotes the average of a specific channel $c$. Finally, we define the loss function of the unsupervised learning scheme as:
\begin{equation}
\label{euq:unsupervision}
\mathcal{L}_{\mathrm{unsup}} =
\mathcal{L}_{\mathrm{D}}
+ \mathcal{L}_{\mathrm{B}}
+ \mathcal{L}_{\mathrm{A}\text{-}\mathrm{unsup}}
+ \lambda_3 \mathcal{L}_{\mathrm{gw}},
\end{equation}
where $\lambda_3$ is a weighting factor.

\textbf{Semi-supervised Learning.} On top of both the bi-directional supervised learning and unsupervised learning schemes, the proposed semi-supervised learning framework can be formulated as:
\begin{equation}
\label{euq:semisupervision}	
\mathcal{L}_{\mathrm{semi}\text{-}\mathrm{sup}} =
\mathcal{L}_{\mathrm{sup}}
+ \lambda_{\mathrm{unsup}} \mathcal{L}_{\mathrm{unsup}},
\end{equation}
where $\lambda_{\mathrm{unsup}}$ is used to balance the two schemes.

\section{Experimental Settings}
\label{experiment}

In this section, we will briefly introduce the baselines, data sets, evaluation metrics and implementation details utilized in our experiments.

\subsection{Baselines}

We compared the proposed method with two prior-based approaches which were developed on top of the IFM \cite{IFM}, including UDCP \cite{UDCP} and DHL \cite{water_hl}, three prior-based methods which did not consider the IFM, including Histogram Equalization (HE) \cite{he}, HLRP \cite{hlrp} and MMLE \cite{mmle}. We also compared our method with seven supervised learning approaches, including WaterNet \cite{waternet}, UColor \cite{ucolor}, FUnIE-GAN \cite{fast_gan}, PUIE-Net (MP) \cite{puie}, U-Transformer \cite{ushape}, URanker \cite{uranker} and CCMSRNet \cite{ccmsrnet}, an unsupervised learning method, i.e., USUIR \cite{usuir}, and three semi-supervised learning methods, including DDFormer \cite{ddformer}, Semi-UIR \cite{semi-uir} and UWFormer \cite{uwformer}.

\subsection{Data Sets}

We conducted a series of experiments using five data sets, including four publicly available real-world underwater data sets, i.e., \textit{UIEB} \cite{waternet}, \textit{SUIM-E} \cite{sguie}, \textit{RUIE} \cite{ruie} and \textit{SQUID} \cite{water_hl}, and a synthetic data set, namely, \textit{EUVP} \cite{fast_gan}. The UIEB data set originally contains 890 pairs of degraded images, in which each reference image was selected from the results produced by applying 12 algorithms to a degraded image. Ten pairs of UIEB \cite{waternet} images were discarded because the degraded images in these pairs had been included in the rest pairs. In total, we only used 880 pairs of UIEB images. 

The SUIM-E \cite{sguie} data set comprises 1,525 pairs of degraded underwater images and a testing set of 110 labeled images. Three subsets are comprised of the RUIE \cite{ruie} data set, including UIQS, UCCS and UHTS, which include 3,630, 300 and 300 unlabeled images, respectively. The SQUID \cite{water_hl} data set comprises 114 underwater images captured using a stereo camera, divided into the Michmoret, Katzaa, Nachsholim and Satil subsets. We utilized 2,185 images from the Underwater Scenes subset of the EUVP \cite{fast_gan} data set. 

Two different PATS-UIENet models were trained for the real-world and synthetic testing images, referred to as $\mathrm{PATS} \text{-} \mathrm{UIENet}_{\mathrm{Real}}$ and $\mathrm{PATS} \text{-} \mathrm{UIENet}_{\mathrm{Syn}}$, respectively. Regarding both the models, we randomly selected 720 pairs from the 880 pairs of UIEB \cite{waternet} images and 2,000 pairs from the Underwater Scenes subset of the EUVP \cite{fast_gan} data set, respectively, which were used as the training images in the supervised learning scheme. We collected 1,934 unlabeled real-world underwater images, which cover diverse underwater scenes, and used these images in the unsupervised learning scheme for both the models. All semi-supervised baselines requiring paired and unlabeled degraded underwater images used the same training settings as those used by PATS-UIENet. Each trainable baseline was also trained with separate UIEB- and EUVP-based models, which were applied to the real-world and synthetic testing sets, respectively. The model was selected according to the data domain, without image-wise selection or fine-tuning.

Among the rest of the UIEB \cite{waternet} data set, 80 pairs of images were randomly selected as the validation set, while the remaining pairs were used as the testing set which is referred to as Test-U80. The UIEB data set also consists of 60 challenging images without references, which were used as the second testing set, referred to as Test-C60. A third testing set was obtained from the testing set of the SUIM-E data set \cite{sguie}, which contains 110 pairs of labeled images and is named as Test-S110. For the RUIE \cite{ruie} data set, the UCCS was utilized as the fourth testing set, denoted as Test-UCCS. Fifty-three SQUID \cite{water_hl} images captured using the camera at the right hand side were randomly selected and were used as the fifth testing set, namely, Test-R53. The remaining 185 images in the Underwater Scenes subset of the EUVP \cite{fast_gan} data set were used as the sixth testing set, referred to as Test-Scenes.

\subsection{Evaluation Metrics}

For the labeled testing data, we used Peak Signal-to-Noise Ratio (PSNR), Structural Similarity Index (SSIM) \cite{ssim} and Learned Perceptual Image Patch Similarity (LPIPS) \cite{lpips} to assess the quality of an enhanced image with regard to the reference image. Regarding the unlabeled testing data, we used the Underwater Image Fidelity (UIF) \cite{uif} and Multi-scale Image Quality (MUSIQ) \cite{musiq} metrics. 
When the SQUID \cite{water_hl} data set was tested, the Average Reproduction Angular Error ($\bar{\psi}$) \cite{water_hl} was used to evaluate the quality of color restoration. Since transmission maps are related to the scene distance, we used the Pearson Correlation Coefficient (PCC) calculated between the estimated transmission maps and the ground-truth depth map to assess the performance of transmission estimation \cite{water_hl}.

\subsection{Implementation Details}
We implemented the proposed method using Pytorch and conducted experiments on Ubuntu 20.04 with a GeForce RTX 3090 graphics processing unit. During the training process, the images were resized to a resolution of $256\times256$ pixels. The number of filters in the RCT was set to 16. For the $Enc\_1$ to $Dec\_1$ in the D-Stream and B-Stream, the numbers of filters were set to 64, 128, 256, 512, 512, 256, 128, 64 and 64 in turn. Regarding the A-Stream, the dimension of each token was 384 and the number of attention heads in the MHSA was 6. We trained the PATS-UIENet using the AdamW \cite{adamw} optimizer. The learning rate and batch size were set to 1e-4 and 12, respectively. Four weighting factors, including $\lambda_1, \lambda_2, \lambda_3$ and $\lambda_{unsup}$, were set to 0.1, 0.005, 1 and 0.1, respectively. We first trained the PATS-UIENet using the bi-directional supervised scheme for 50 epochs as a warm-up stage. Then the semi-supervised scheme was used to train it for 1000 epochs. When the semi-supervised learning process was carried out, in particular, the unsupervised scheme was performed for 30 iterations after the bi-directional supervised scheme was conducted for each epoch.

\section{Experimental Results}
\label{results}

In this section, we will report the results obtained in the underwater image enhancement, transmission estimation and color restoration experiments and the ablation study.

\subsection{Underwater Image Enhancement}
We evaluated the proposed PATS-UIENet along with sixteen baselines using three full-reference quantitative metrics and two non-reference quantitative metrics. A qualitative analysis and a performance analysis were also performed. In this subsection, the results obtained in the four experiments will be reported.

\subsubsection{Full-Reference Quantitative Evaluation} Since the Test-U80, Test-S110 and Test-Scenes testing sets contain reference images, a full-reference quantitative evaluation was conducted on the proposed method and the 16 baselines using the PSNR, SSIM \cite{ssim} and LPIPS \cite{lpips} metrics across these testing sets. The results are shown in Table \ref{tab:ref}. As can be seen, our PATS-UIENet normally achieved the best performance across all three metrics, regardless of the testing set. Compared with the baselines designed on top of the IFM \cite{IFM}, such as UDCP \cite{UDCP} and DHL \cite{hl}, and the existing semi-supervised learning approaches, including DDFormer \cite{ddformer}, Semi-UIR \cite{semi-uir} and UWFormer \cite{uwformer}, our method normally showed a large margin.

\begin{table}[t]
\centering
\caption{Full-reference quantitative evaluation of the proposed PATS-UIENet and sixteen baselines on three real-world test sets. The best and second-best results are highlighted in \textbf{bold} and \underline{underline}, respectively.}
\label{tab:ref}
{\fontsize{8pt}{8pt}\selectfont
\setlength{\tabcolsep}{0.5mm}
\begin{tabular}{cccccccccc}
\toprule
\multirow{2}{*}{Method} & \multicolumn{3}{c}{Test-U80} & \multicolumn{3}{c}{Test-S110} & \multicolumn{3}{c}{Test-Scenes} \\
\cmidrule{2-10}
 & PSNR$\uparrow$ & SSIM$\uparrow$ & LPIPS$\downarrow$ 
 & PSNR$\uparrow$ & SSIM$\uparrow$ & LPIPS$\downarrow$ 
 & PSNR$\uparrow$ & SSIM$\uparrow$ & LPIPS$\downarrow$ \\
\midrule

UDCP \cite{UDCP} & 9.51 & 33.66 & 41.74 & 10.07 & 34.29 & 37.66 & 14.76 & 56.43 & 30.31 \\
DHL \cite{water_hl} & 15.16 & 63.93 & 30.40 & 14.61 & 62.93 & 28.39 & 14.99 & 69.73 & 22.36 \\
HE \cite{he} & 16.60 & 77.86 & 25.67 & 15.41 & 74.79 & 30.05 & 13.61 & 62.82 & 36.10 \\
HLRP \cite{hlrp} & 13.56 & 22.76 & 33.90 & 12.55 & 29.24 & 33.18 & 12.12 & 18.21 & 39.35 \\
MMLE \cite{mmle} & 18.56 & 76.21 & 22.57 & 17.32 & 77.39 & 22.77 & 14.89 & 62.23 & 31.99 \\

\midrule

WaterNet \cite{waternet} & 17.13 & 70.41 & 40.01 & 18.83 & 75.22 & 31.08 & \underline{25.84} & \underline{84.02} & 16.28 \\
UColor \cite{ucolor} & 21.05 & 85.24 & 17.97 & 20.26 & 84.25 & 15.07 & 24.31 & 81.40 & 19.05 \\
FUnIE-GAN \cite{fast_gan} & 15.31 & 59.68 & 44.28 & 16.89 & 66.32 & 34.98 & 25.11 & 83.16 & \underline{13.76} \\
USUIR \cite{usuir} & 16.92 & 68.69 & 28.63 & 17.54 & 76.58 & 21.38 & 15.14 & 66.51 & 32.44 \\
PUIE-Net (MP) \cite{puie} & 22.26 & \underline{88.89} & \underline{12.01} & 21.85 & 89.19 & \underline{9.70} & 21.85 & 73.47 & 24.45 \\
U-Transformer \cite{ushape} & 20.98 & 72.76 & 26.32 & 19.92 & 68.25 & 30.17 & 24.09 & 79.98 & 15.08 \\
URanker \cite{uranker} & 21.87 & 85.75 & 18.64 & 21.27 & 87.43 & 13.51 & 22.46 & 82.35 & 20.87 \\
CCMSRNet \cite{ccmsrnet} & 22.74 & 88.74 & 13.18 & \underline{22.22} & 89.07 & 10.76 & 18.96 & 77.67 & 23.85 \\
DDFormer \cite{ddformer} & 10.94 & 25.74 & 58.26 & 10.87 & 41.27 & 72.19 & 10.64 & 27.81 & 69.46 \\
Semi-UIR \cite{semi-uir} & \textbf{23.63} & 81.81 & 23.08 & 19.83 & 73.95 & 30.54 & 19.12 & 74.94 & 25.68 \\
UWFormer \cite{uwformer} & 19.65 & 85.26 & 17.21 & 21.45 & \underline{90.27} & 10.88 & 19.62 & 80.36 & 21.05 \\

\midrule

PATS-UIENet (Ours) & \underline{23.59} & \textbf{90.16} & \textbf{10.42} & \textbf{22.78} & \textbf{90.54} & \textbf{8.85} & \textbf{25.87} & \textbf{84.34} & \textbf{12.04} \\

\bottomrule
\end{tabular}
}

\end{table}

\subsubsection{Non-reference Quantitative Evaluation} We used two non-reference metrics, i.e., UIF \cite{uif} and MUSIQ \cite{musiq}, to assess the performance of our method and sixteen baselines on the Test-U80, Test-S110, Test-Scenes, Test-C60, Test-UCCS and Test-R53 testing sets. The results are shown in Tables~\ref{tab:nonref_part1} and \ref{tab:nonref_part2}. It can be seen that our method achieved the best UIF score on three out of the six testing sets, including Test-U80, Test-S110 and Test-C60, and ranked the second on the Test-R53 testing set. Regarding the MUSIQ metric, the proposed method still demonstrated the competitive performance, especially on the challenging Test-S110, Test-Scenes and Test-R53 testing sets, even though it did not always produce the highest score.

\begin{table}[t]
\centering
\caption{Non-reference quantitative evaluation of the proposed PATS-UIENet and sixteen baselines on Test-U80, Test-S110 and Test-Scenes. The best and second-best results are highlighted in \textbf{bold} and \underline{underline}, respectively.}
\label{tab:nonref_part1}
{\fontsize{8pt}{8pt}\selectfont
\setlength{\tabcolsep}{2.5mm}
\begin{tabular}{cccccccc}
\toprule
\multirow{2}{*}{Method} 
& \multicolumn{2}{c}{Test-U80} 
& \multicolumn{2}{c}{Test-S110} 
& \multicolumn{2}{c}{Test-Scenes} \\
\cmidrule{2-7}
& UIF$\uparrow$ & MUSIQ$\uparrow$ 
& UIF$\uparrow$ & MUSIQ$\uparrow$ 
& UIF$\uparrow$ & MUSIQ$\uparrow$ \\
\midrule

UDCP & 36.40 & 44.99 & 32.01 & 51.38 & 61.25 & 36.99 \\
DHL & 41.32 & 47.43 & 14.80 & 57.22 & 0.55 & 36.57 \\
HE & 44.04 & 46.68 & 45.01 & 56.14 & 52.81 & 35.79 \\
HLRP & 0.59 & 49.73 & 0.91 & 56.87 & 0.71 & 40.21 \\
MMLE & 30.99 & \textbf{52.50} & 35.05 & 60.18 & 42.05 & 44.67 \\

\midrule

WaterNet & 33.25 & 31.84 & 28.50 & 40.47 & \textbf{72.94} & 38.47 \\
UColor & 44.15 & 45.10 & 42.83 & 54.80 & 70.20 & 35.12 \\
FUnIE-GAN & 29.12 & 47.47 & 22.77 & 52.27 & 37.69 & 45.38 \\
USUIR & 5.17 & 44.44 & 37.02 & 56.83 & 48.82 & 37.91 \\
PUIE-Net (MP) & 9.18 & 49.05 & \underline{60.98} & 60.44 & 27.69 & \underline{45.46} \\
U-Transformer & 5.16 & 39.34 & 24.94 & 44.96 & 38.68 & 30.44 \\
URanker & \underline{54.87} & 44.63 & 58.31 & 57.15 & \underline{72.54} & 39.20 \\
CCMSRNet & 54.61 & 49.85 & 0.23 & 60.59 & 63.59 & 39.64 \\
DDFormer & -5.36 & 36.99 & 1.92 & 40.68 & -3.16 & 23.87 \\
Semi-UIR & 50.06 & 45.40 & 31.94 & 51.22 & 44.01 & 41.47 \\
UWFormer & 46.48 & \underline{50.95} & 55.69 & \textbf{61.53} & 67.40 & 36.80 \\

\midrule

PATS-UIENet (Ours) & \textbf{58.83} & 49.91 & \textbf{61.20} & \underline{60.66} & 68.07 & \textbf{46.70} \\

\bottomrule
\end{tabular}
}
\end{table}

\begin{table}[t]
\centering
\caption{Non-reference quantitative evaluation of the proposed PATS-UIENet and sixteen baselines on Test-C60, Test-UCCS and Test-R53. The best and second-best results are highlighted in \textbf{bold} and \underline{underline}, respectively.}
\label{tab:nonref_part2}
{\fontsize{8pt}{8pt}\selectfont
\setlength{\tabcolsep}{2.5mm}
\begin{tabular}{cccccccc}
\toprule
\multirow{2}{*}{Method} 
& \multicolumn{2}{c}{Test-C60} 
& \multicolumn{2}{c}{Test-UCCS} 
& \multicolumn{2}{c}{Test-R53} \\
\cmidrule{2-7}
& UIF$\uparrow$ & MUSIQ$\uparrow$ 
& UIF$\uparrow$ & MUSIQ$\uparrow$ 
& UIF$\uparrow$ & MUSIQ$\uparrow$ \\
\midrule

UDCP & 25.41 & 33.03 & 34.72 & 29.72 & 3.24 & 36.85 \\
DHL & 25.04 & 37.27 & 3.62 & 31.07 & 2.53 & 42.64 \\
HE & 41.50 & 36.08 & 38.08 & 31.24 & 1.67 & 43.06 \\
HLRP & 3.57 & 34.65 & 0.32 & 34.44 & 1.26 & 43.53 \\
MMLE & 24.95 & 40.18 & 28.08 & \textbf{35.69} & 24.67 & \textbf{53.53} \\

\midrule

WaterNet & 28.65 & 34.75 & 33.38 & 24.88 & 9.71 & 32.93 \\
UColor & 26.46 & 36.01 & 49.22 & \underline{35.12} & 2.31 & \underline{49.63} \\
FUnIE-GAN & 23.11 & \textbf{43.49} & 22.67 & 34.64 & 10.93 & 38.92 \\
USUIR & 22.35 & 34.15 & 20.54 & 28.83 & 17.98 & 39.44 \\
PUIE-Net (MP) & 3.39 & 38.69 & \underline{51.76} & 29.18 & \textbf{46.16} & 45.15 \\
U-Transformer & 29.31 & 37.66 & 29.69 & 27.05 & 19.24 & 30.06 \\
URanker & 44.70 & 38.37 & 45.97 & 29.60 & 39.19 & 41.92 \\
CCMSRNet & 41.85 & 40.24 & 51.34 & 33.72 & 34.80 & 48.23 \\
DDFormer & 11.14 & 36.21 & -2.14 & 20.19 & 5.86 & 24.87 \\
Semi-UIR & \underline{46.82} & \underline{43.25} & 32.28 & 30.34 & 23.33 & 35.21 \\
UWFormer & 36.76 & 41.40 & \textbf{53.24} & 34.30 & 39.98 & 46.26 \\

\midrule

PATS-UIENet (Ours) & \textbf{46.99} & 40.49 & 51.18 & 31.63 & \underline{42.51} & 46.63 \\

\bottomrule
\end{tabular}
}
\end{table}

\subsubsection{Qualitative Analysis} 
\begin{figure}[t]
\centering    \includegraphics[width=0.99\textwidth]{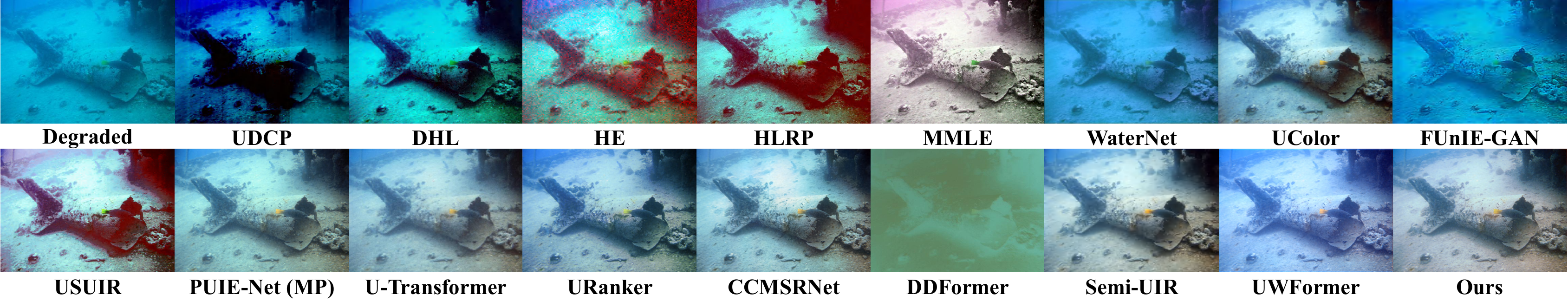}
\caption{The results produced by 16 baselines and our method in terms of a degraded image in the Test-U80 testing set.}
\label{fig:results_u80}
\end{figure}

\begin{figure}[t]
\centering    \includegraphics[width=0.99\textwidth]{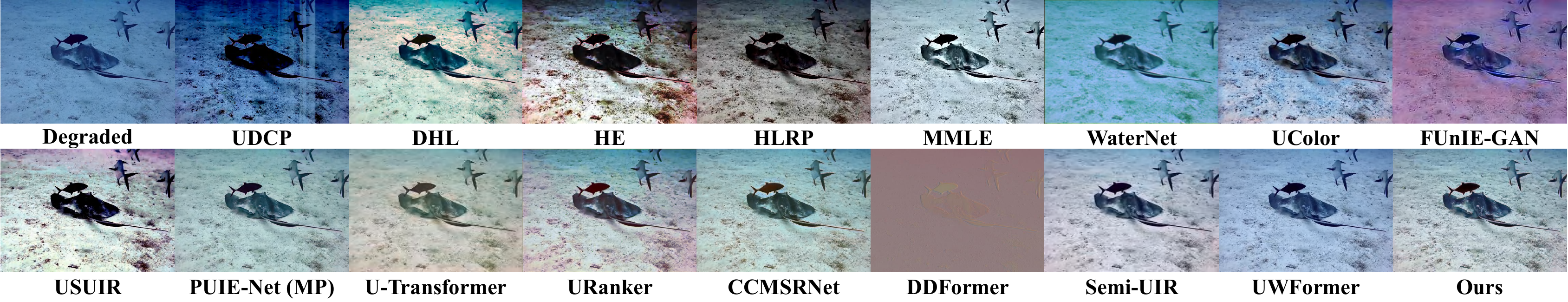}
 \caption{The results produced by 16 baselines and our method in terms of a degraded image in the Test-S110 testing set.}
\label{fig:results_s110}
\end{figure}

\begin{figure}[t]
\centering    \includegraphics[width=0.99\textwidth]{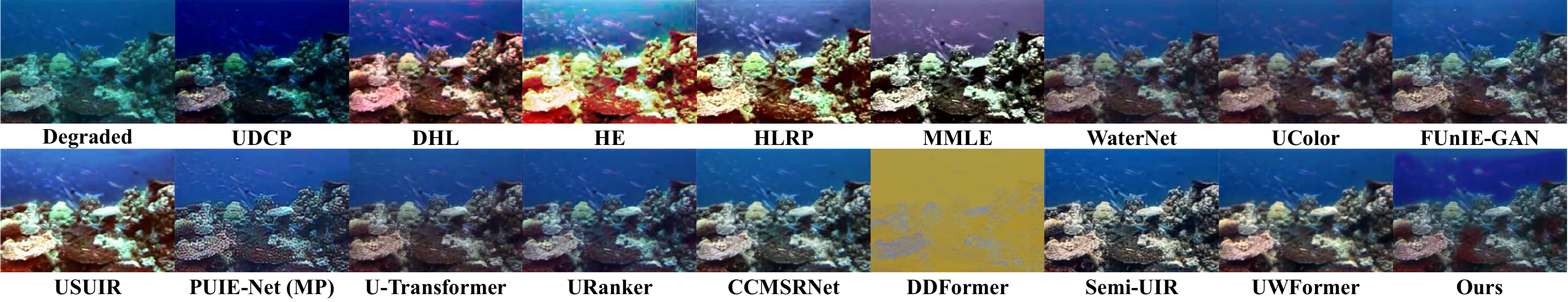}
 \caption{The results produced by 16 baselines and our method in terms of a degraded image in the Test-Scenes testing set.}
\label{fig:results_scene}
\end{figure}

\begin{figure}[t]
\centering    \includegraphics[width=0.99\textwidth]{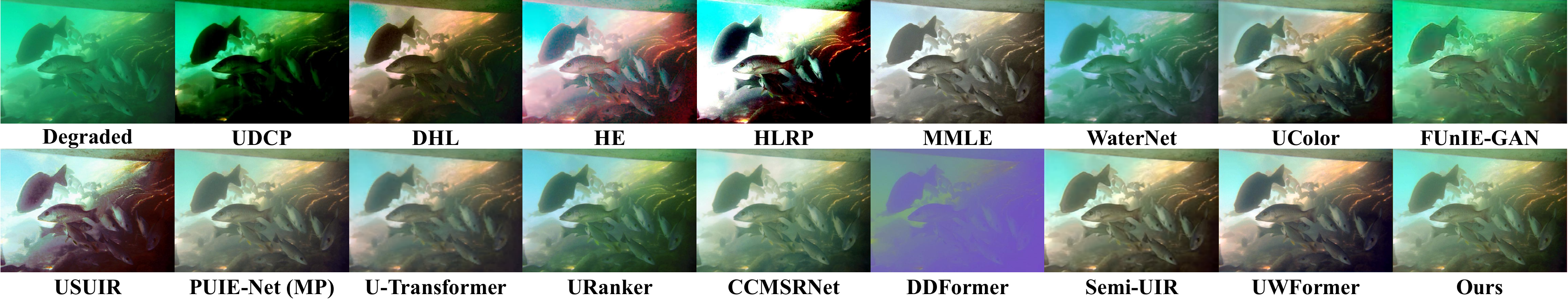}
 \caption{The results produced by 16 baselines and our method in terms of a degraded image in the Test-C60 testing set.}
\label{fig:results_c60}
\end{figure}

\begin{figure}[t]
\centering    \includegraphics[width=0.99\textwidth]{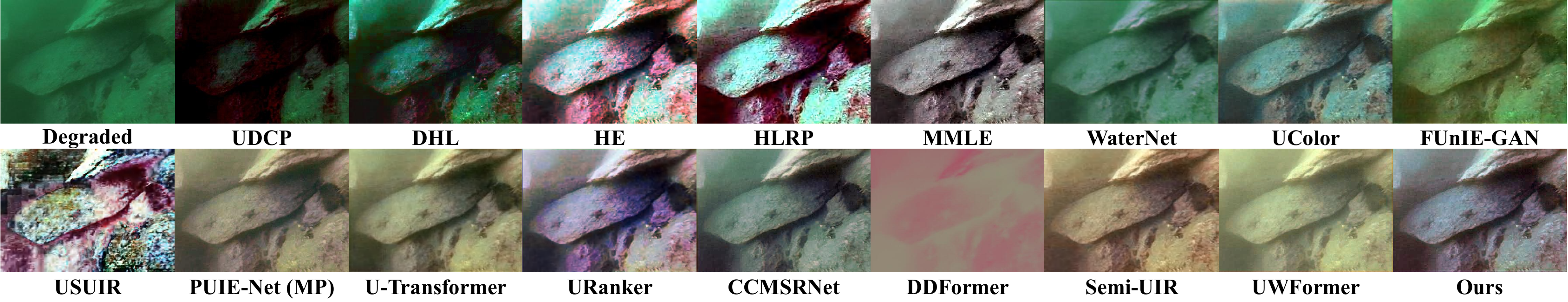}
 \caption{The results produced by 16 baselines and our method in terms of a degraded image in the Test-UCCS testing set.}
\label{fig:results_uccs}
\end{figure}

\begin{figure}[t]
\centering    \includegraphics[width=0.99\textwidth]{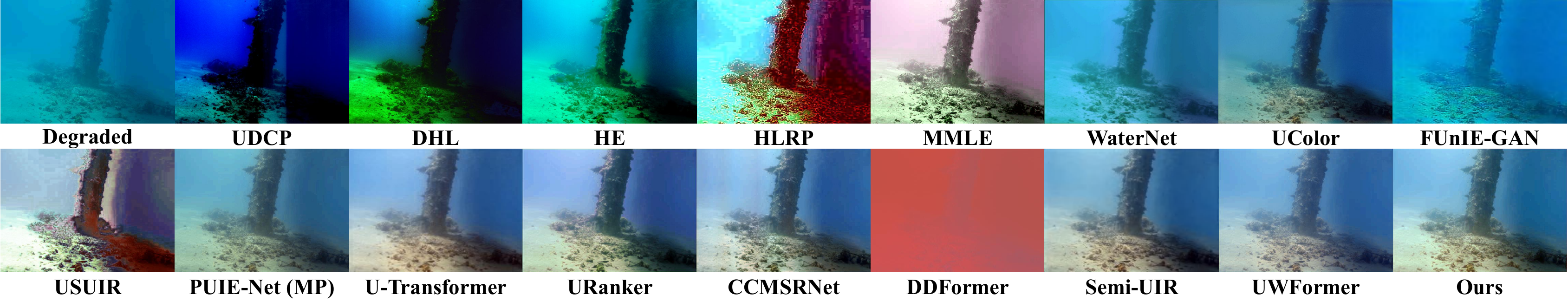}
\caption{The results produced by 16 baselines and our method in terms of a degraded image in the Test-R53 testing set.}
\label{fig:results_r53}
\end{figure}

The enhanced images generated by the 16 baselines and our method on six testing sets are shown in Figs. \ref{fig:results_u80} - \ref{fig:results_r53}, respectively. It can be seen that some methods with the higher UIF \cite{uif} or MUSIQ \cite{musiq} score did not produce visually pleasing results. For example, MMLE \cite{mmle} produced the highest MUSIQ score on Test-U80 and Test-R53, but the resulting images suffered from pale colors and over-enhancement artifacts (see Figs. \ref{fig:results_u80} and \ref{fig:results_r53}). In contrast, our PATS-UIENet improved different degraded images and produced images with the natural and vivid color. Although the UIF \cite{uif} or MUSIQ \cite{musiq} scores that our method produced were slightly lower than those obtained using some baseline methods \cite{mmle, waternet, fast_gan} on certain testing sets, the enhanced images still manifest visually satisfactory quality.

\subsubsection{Performance Analysis}
We compared the proposed PATS-UIENet with 11 deep baseline methods in terms of the number of parameters, Floating Point Operations Per Second (FLOPs) and inference speed (ms). Both the number of parameters and the FLOPs were calculated using the \textit{ptflops}\footnote{\url{https://pypi.org/project/ptflops/}} tool. As shown in Table~\ref{tab:complexity}, the values of number of parameters, FLOPs and inference speed vary significantly. Specifically, UColor \cite{ucolor} exhibits the largest number of parameters (148.04M) and the largest FLOPs value (2804.36G), while UWFormer \cite{uwformer} incurs the slowest inference speed (491.00 ms). On the other hand, USUIR \cite{usuir} has the lightest design with only 0.23M parameters and has a FLOPs value of 29.62G and owns the fastest inference speed (2.13ms). In contrast, our PATS-UIENet achieved a proper balance between the number of parameters or FLOPs and the inference speed.

\begin{table}[t]
\centering
\small
\caption{Comparison among 11 deep UIE methods and the proposed PATS-UIENet in terms of the number of parameters, FLOPs, and inference speed.}
\label{tab:complexity}
{\fontsize{8pt}{8pt}\selectfont
\setlength{\tabcolsep}{5mm}

\begin{tabular}{cccc}
\toprule
Method & \#Params (M) & FLOPs (G) & Speed (ms) \\
\midrule

WaterNet \cite{waternet} & 1.09 & 142.84 & 6.40 \\
UColor \cite{ucolor} & 148.04 & 2804.36 & 15.51 \\
FUnIE-GAN \cite{fast_gan} & 7.02 & 20.48 & 73.18 \\
USUIR \cite{usuir} & 0.23 & 29.62 & 2.13 \\
PUIE-Net (MP) \cite{puie} & 1.40 & 70.54 & 101.89 \\
U-Transformer \cite{ushape} & 22.80 & 5.96 & 91.30 \\
URanker \cite{uranker} & 3.15 & 20.90 & 45.50 \\
CCMSRNet \cite{ccmsrnet} & 21.69 & 87.18 & 152.51 \\
DDFormer \cite{ddformer} & 7.63 & 35.54 & 50.04 \\
Semi-UIR \cite{semi-uir} & 1.65 & 72.88 & 84.93 \\
UWFormer \cite{uwformer} & 29.84 & 30.18 & 491.00 \\

\midrule
PATS-UIENet (Ours) & 45.71 & 179.70 & 86.17 \\
\bottomrule

\end{tabular}
}
\end{table}

\subsection{Transmission Estimation and Color Restoration}

\begin{figure*}[t]
\centering    
\includegraphics[width=0.99\textwidth]{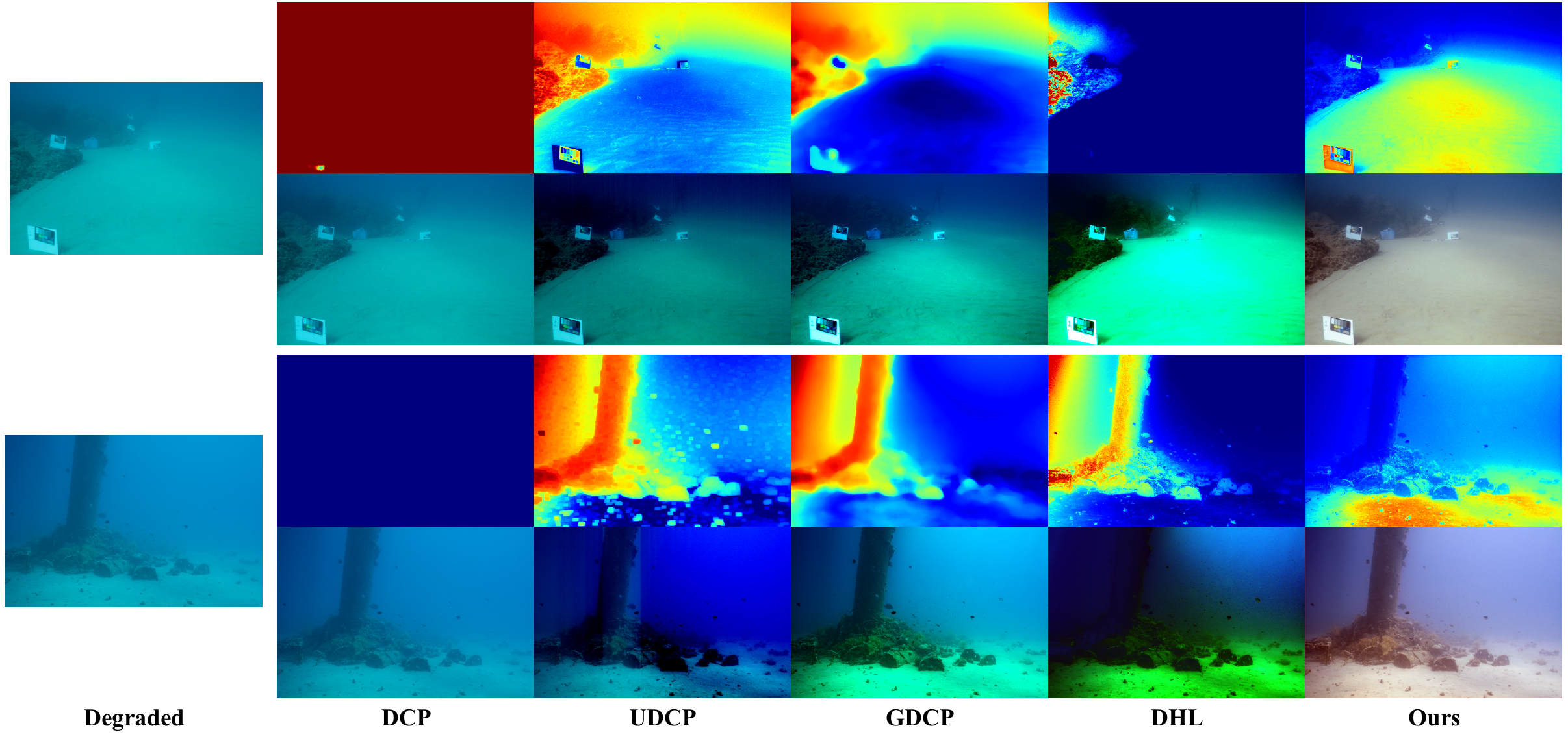}
\caption{Each group shows a degraded image and the red channels of five transmission maps (top) and five enhanced images (bottom) obtained using four prior-based baselines and our method.}
\label{fig:estimation}
\end{figure*}

The quantitative evaluation of our methods and four prior-based baselines for transmission estimation and color restoration is reported in Table \ref{tab:squid}. It can be seen that our method achieved the superior, or at least the comparable, performance to that of the four prior-based baselines on four different subsets of Test-R53. Specifically, our method outperformed the baselines with a large margin on both the Michmoret and Katzaa subsets, no matter which metric was considered. 
It should be noted that none of these methods produced a PCC value higher than 0.12 on the Satil subset. The inferior results may be attributed to the special scenes contained in this subset \cite{water_hl_2}. As shown in Fig. \ref{fig:estimation}, our method was able to perform transmission estimation and color restoration by directly learning from the limited real-world underwater images.

\begin{table}[t]
\centering
\small
\caption{Quantitative evaluation of four prior-based baselines and the proposed method on four subsets of Test-R53, where PCC and $\bar{\psi}$ are used for transmission estimation and color restoration, respectively. The best and second-best results are highlighted in \textbf{bold} and \underline{underline}, respectively.}
\label{tab:squid}
{\fontsize{8pt}{8pt}\selectfont
\setlength{\tabcolsep}{2.5mm}
\begin{tabular}{ccccccccc}
\toprule
\multirow{2}{*}{Method} 
& \multicolumn{2}{c}{Michmoret} 
& \multicolumn{2}{c}{Katzaa} 
& \multicolumn{2}{c}{Nachsholim} 
& \multicolumn{2}{c}{Satil} \\
\cmidrule{2-9}
& PCC$\uparrow$ & $\bar{\psi}\downarrow$ 
& PCC$\uparrow$ & $\bar{\psi}\downarrow$ 
& PCC$\uparrow$ & $\bar{\psi}\downarrow$ 
& PCC$\uparrow$ & $\bar{\psi}\downarrow$ \\
\midrule

DCP \cite{DCP} & -0.16 & 34.63 & 0.00 & 35.47 & -0.26 & 34.75 & 0.03 & 36.21 \\
UDCP \cite{UDCP} & -0.54 & 36.97 & -0.15 & 40.43 & \underline{0.06} & 38.80 & \textbf{0.12} & 51.45 \\
GDCP \cite{GDCP} & \underline{0.29} & 33.76 & \underline{0.24} & \underline{34.47} & -0.04 & 34.28 & -0.03 & 35.31 \\
DHL \cite{water_hl} & -0.07 & \underline{32.10} & 0.10 & 35.50 & \textbf{0.36} & 35.92 & \underline{0.11} & 34.82 \\

\midrule  

Ours & \textbf{0.62} & \textbf{12.83} & \textbf{0.32} & \textbf{9.47} & -0.28 & \textbf{15.31} & 0.04 & \textbf{13.14} \\

\bottomrule
\end{tabular}
}
\end{table}

\subsection{Ablation Study}
To investigate the effect of different components of the PATS-UIENet, we conducted a series of ablation experiments. For simplicity, we only utilized Test-U80 and the Michmorest subset in Test-R53.

\subsubsection{Effect of the Semi-supervised Learning Framework}
To validate the effect of the proposed semi-supervised learning framework, we compared it with the unsupervised learning, supervised learning and bi-directional supervised learning frameworks. As reported in Table~\ref{tab:ablation_learning}, the proposed semi-supervised framework always outperformed the other frameworks in terms of different metrics across different testing sets. It has been suggested that our network can be better trained using the semi-supervised learning framework with both the labeled and unlabeled real-world images, which improves the enhancement performance.

\begin{table}[t]
\centering
\small
\caption{Comparison of different learning frameworks on Test-U80 and the Michmoret subset of Test-R53. The best and second-best results are highlighted in \textbf{bold} and \underline{underline}, respectively.}
\label{tab:ablation_learning}
{\fontsize{8pt}{8pt}\selectfont
\setlength{\tabcolsep}{3.8mm}
\begin{tabular}{cccccc}
\toprule
\multirow{2}{*}{\makecell{Learning\\Framework}} 
& \multicolumn{3}{c}{Test-U80} 
& \multicolumn{2}{c}{Michmoret} \\
\cmidrule{2-6}
& PSNR$\uparrow$ & SSIM$\uparrow$ & LPIPS$\downarrow$ 
& PCC$\uparrow$ & $\bar{\psi}\downarrow$ \\
\midrule

Unsupervised & 16.85 & 78.44 & 29.91 & 0.50 & 15.46 \\
Supervised & 22.19 & 88.23 & 13.07 & 0.57 & 17.01 \\
Bi-supervised & \underline{23.00} & \underline{89.49} & \underline{11.93} & \underline{0.61} & \underline{15.02} \\

\midrule

Semi-supervised (Ours) & \textbf{23.59} & \textbf{90.16} & \textbf{10.42} & \textbf{0.62} & \textbf{12.83} \\

\bottomrule
\end{tabular}
}
\end{table}

\subsubsection{Effect of the RCT and RCM} 
To examine the effect of the RCT and RCM on the performance of our PATS-UIENet, we removed the RCT, the RCM or both of them. In this case, we derived three variants of the PATS-UIENet, including one with both the RCT and RCM removed, one with only the RCT removed and one with only the RCM removed, denoted as ``Simplified", ``w/o RCT" and ``w/o RCM", respectively. To verify the optimality of applying the RCT, we further replaced the RCT with a Blue Channel Tuner (BCT) and a Green Channel Tuner (GCT) separately, which tune the blue and green channels, respectively, instead of the red channel. The two variants are denoted as ``w/ BCT'' and ``w/ GCT'', respectively. As shown in Table~\ref{tab:ablation_module}, the complete PATS-UIENet achieved the best performance in terms of all the metrics. Removal of the RCT led to a slight increase in the PSNR and SSIM metrics compared to the Simplified version, but worsened the LPIPS and $\bar{\psi}$ scores. 

On the other hand, replacing the RCT with the BCT or GCT improved the performance compared with the Simplified version, indicating that channel-wise tuning is beneficial for underwater image enhancement. However, both the BCT and GCT variants were still inferior to the complete model with the RCT across all metrics. This finding demonstrates that tuning the red channel is more effective than tuning the blue or green channel, which is consistent with the physical observation that red light suffers from more severe attenuation in underwater environments.

In addition, removal of the RCM decreased the performance with regard to all metrics across the two data sets, confirming its usefulness in facilitating information exchange between the B-Stream and A-Stream. The results indicate that both the RCT and RCM contribute meaningfully to the performance of our method.

\begin{table}[t]
\centering
\small
\caption{Impact of the RCT and RCM modules on the performance of the proposed method on Test-U80 and the Michmoret subset of Test-R53. The best and second-best results are highlighted in \textbf{bold} and \underline{underline}, respectively.}
\label{tab:ablation_module}
{\fontsize{8pt}{8pt}\selectfont
\setlength{\tabcolsep}{5mm}
\begin{tabular}{cccccc}
\toprule
\multirow{2}{*}{Variant} 
& \multicolumn{3}{c}{Test-U80} 
& \multicolumn{2}{c}{Michmoret} \\
\cmidrule{2-6}
& PSNR$\uparrow$ & SSIM$\uparrow$ & LPIPS$\downarrow$ 
& PCC$\uparrow$ & $\bar{\psi}\downarrow$ \\
\midrule

Simplified & 22.96 & 89.64 & 11.25 & 0.59 & 13.59 \\
w/o RCM & 22.10 & 88.83 & 12.20 & 0.60 & 13.50 \\
w/o RCT & 23.14 & \underline{90.03} & 12.20 & 0.60 & 13.97 \\
w/ BCT & \underline{23.45} & 89.92 & \underline{10.58} & \underline{0.61} & 13.04 \\
w/ GCT & 23.06 & 89.87 & 10.78 & 0.60 & \underline{12.92} \\

\midrule

Ours & \textbf{23.59} & \textbf{90.16} & \textbf{10.42} & \textbf{0.62} & \textbf{12.83} \\

\bottomrule
\end{tabular}
}
\end{table}

\subsubsection{Effect of the Degradation Control Factor} 
To evaluate the impact of the degradation control factor \(\alpha\) on the performance of the PATS-UIENet, we conducted an ablation experiment on testing three different values of \(\alpha\). As shown in Table~\ref{tab:ablation_hyperparameter1}, the default setting of \(\alpha\) = 0.1 achieved the best performance with regard to both the full-reference and the non-reference metrics across the two data sets.

\begin{table}[t]
\centering
\small
\caption{Effect of the degradation control factor $\alpha$ on the performance of the proposed method. The best and second-best results are highlighted in \textbf{bold} and \underline{underline}, respectively.}
\label{tab:ablation_hyperparameter1}
{\fontsize{8pt}{8pt}\selectfont
\setlength{\tabcolsep}{5mm}
\begin{tabular}{cccccc}
\toprule
\multirow{2}{*}{$\alpha$} 
& \multicolumn{3}{c}{Test-U80} 
& \multicolumn{2}{c}{Michmoret} \\
\cmidrule{2-6}
& PSNR$\uparrow$ & SSIM$\uparrow$ & LPIPS$\downarrow$ 
& PCC$\uparrow$ & $\bar{\psi}\downarrow$ \\
\midrule

0.001 & \underline{23.21} & \underline{89.77} & \underline{11.02} & 0.59 & \underline{13.92} \\
1 & 23.06 & 89.72 & 11.08 & \underline{0.61} & 14.11 \\  

\midrule

0.1 (Ours) & \textbf{23.59} & \textbf{90.16} & \textbf{10.42} & \textbf{0.62} & \textbf{12.83} \\ 

\bottomrule
\end{tabular}
}

\end{table}

\subsubsection{Effect of Loss Hyperparameters} 
To evaluate the impact of the hyperparameters used for the loss functions, including \(\lambda_1\), \(\lambda_2\), \(\lambda_3\) and \(\lambda_{\mathrm{unsup}}\), on the performance of the proposed PATS-UIENet, we conducted a comprehensive ablation study by changing their values. Regarding the supervised learning loss function (see Eq. (\ref{euq:bi-supervision})), the results produced by our method with different combinations of the $\lambda_1$ and $\lambda_2$ values are shown in Table~\ref{tab:ablation_hyperparameter2}. It can be seen that the combination that we used produced the best result in terms of each metric. For the unsupervised learning loss function (see Eq. (\ref{euq:unsupervision})), the results produced by our method with different values of $\lambda_3$ are reported in Table~\ref{tab:ablation_hyperparameter3}. Again, the value that we chose led to the best result no matter what metric was considered. In terms of the semi-supervised learning loss function (see Eq. (\ref{euq:semisupervision})), we present the results obtained using our method with different $\lambda_{\mathrm{unsup}}$ values in Table~\ref{tab:ablation_hyperparameter4}. As can be observed, the value that we utilized produced the best result with regard to each metric. The above findings highlight the effectiveness of the values of different loss hyperparameters that we chose.

\begin{table}[t]
\centering
\small
\caption{Effect of different combinations of hyperparameters used in the supervised loss function, i.e., $\lambda_1$ and $\lambda_2$, on the performance of the proposed method. The best and second-best results are highlighted in \textbf{bold} and \underline{underline}, respectively.}
\label{tab:ablation_hyperparameter2}
{\fontsize{8pt}{8pt}\selectfont
\setlength{\tabcolsep}{3.5mm}
\begin{tabular}{ccccccc}
\toprule
\multirow{2}{*}{$\lambda_1$} & \multirow{2}{*}{$\lambda_2$} 
& \multicolumn{3}{c}{Test-U80} 
& \multicolumn{2}{c}{Michmoret} \\
\cmidrule{3-7}
& & PSNR$\uparrow$ & SSIM$\uparrow$ & LPIPS$\downarrow$ 
& PCC$\uparrow$ & $\bar{\psi}\downarrow$ \\
\midrule

1 & 0.0005 & 22.04 & 88.01 & 12.94 & 0.36 & 16.92 \\
0.0001 & 0.0005 & 22.84 & 89.65 & \underline{10.85} & \underline{0.61} & \underline{13.57} \\ 
0.1 & 0.1 & \underline{22.95} & \underline{89.67} & 11.52 & \underline{0.61} & 15.69 \\
0.1 & 0 & 22.91 & 80.57 & 11.54 & 0.59 & 16.92 \\

\midrule

0.1 & 0.0005 (Ours) & \textbf{23.59} & \textbf{90.16} & \textbf{10.42} & \textbf{0.62} & \textbf{12.83} \\

\bottomrule
\end{tabular}
}

\end{table}

\begin{table}[t]
\centering
\small
\caption{Effect of the weighting factor $\lambda_3$ in the unsupervised loss function on the performance of the proposed method. The best and second-best results are highlighted in \textbf{bold} and \underline{underline}, respectively.}
\label{tab:ablation_hyperparameter3}
{\fontsize{8pt}{8pt}\selectfont
\setlength{\tabcolsep}{5.5mm}
\begin{tabular}{cccccc}
\toprule
\multirow{2}{*}{$\lambda_3$} 
& \multicolumn{3}{c}{Test-U80} 
& \multicolumn{2}{c}{Michmoret} \\
\cmidrule{2-6}
& PSNR$\uparrow$ & SSIM$\uparrow$ & LPIPS$\downarrow$ 
& PCC$\uparrow$ & $\bar{\psi}\downarrow$ \\
\midrule

0.1 & \underline{22.95} & \underline{89.80} & \underline{11.01} & \underline{0.61} & 15.14 \\
10 & 22.36 & 88.19 & 13.09 & 0.60 & \underline{14.72} \\  

\midrule

1 (Ours) & \textbf{23.59} & \textbf{90.16} & \textbf{10.42} & \textbf{0.62} & \textbf{12.83} \\ 

\bottomrule
\end{tabular}
}

\end{table}

\begin{table}[t]
\centering
\small
\caption{Effect of the weighting factor $\lambda_{\mathrm{unsup}}$ in the semi-supervised learning scheme on the performance of the proposed method. The best and second-best results are highlighted in \textbf{bold} and \underline{underline}, respectively.}
\label{tab:ablation_hyperparameter4}
{\fontsize{8pt}{8pt}\selectfont
\setlength{\tabcolsep}{5.3mm}
\begin{tabular}{cccccc}
\toprule
\multirow{2}{*}{$\lambda_{\mathrm{unsup}}$} 
& \multicolumn{3}{c}{Test-U80} 
& \multicolumn{2}{c}{Michmoret} \\
\cmidrule{2-6}
& PSNR$\uparrow$ & SSIM$\uparrow$ & LPIPS$\downarrow$ 
& PCC$\uparrow$ & $\bar{\psi}\downarrow$ \\
\midrule

1 & 22.32 & 88.75 & 12.16 & \underline{0.61} & 14.08 \\
0.01 & \underline{23.31} & \underline{89.96} & \underline{10.92} & \underline{0.61} & \underline{12.84} \\ 

\midrule

0.1 (Ours) & \textbf{23.59} & \textbf{90.16} & \textbf{10.42} & \textbf{0.62} & \textbf{12.83} \\ 

\bottomrule
\end{tabular}
}

\end{table}

\subsubsection{Effect of Different Stream Architectures} 
To further examine the effect of different stream architectures on the PATS-UIENet, we constructed its three variants by building the D-Stream, B-Stream and A-Stream using convolution and/or Transformer networks. In addition, we obtained a fourth variant by removing the D-Stream. In essence, this variant is equal to the network built on top of the original IFM \cite{IFM}. The four variants were trained and tested using the same setup as that used for our PATS-UIENet. The results produced by these variants and our method are shown in Table~\ref{tab:ablation_streams1}. It can be seen that our method produced the better, or at least comparable, results in contrast to the four variants. This finding supports the design philosophy of the PATS-UIENet. That is to say, CNNs are suitable for capturing local characteristics while Transformer is good at encoding global characteristics. Moreover, our network which was built on top of the revised IFM \cite{revised_IFM} normally outperformed the variant constructed based on the original IFM \cite{IFM}, confirming the choice of the revised IFM \cite{revised_IFM}.

\begin{table}[t]
\centering
\small
\caption{Comparison between the proposed method and four variants constructed by using different networks for the D-, B-, and A-streams. The best and second-best results are highlighted in \textbf{bold} and \underline{underline}, respectively.}
\label{tab:ablation_streams1}
{\fontsize{8pt}{8pt}\selectfont
\setlength{\tabcolsep}{2.3mm}
\begin{tabular}{ccccccccc}
\toprule
\multirow{2}{*}{Method} 
& \multicolumn{3}{c}{Streams} 
& \multicolumn{3}{c}{Test-U80} 
& \multicolumn{2}{c}{Michmoret} \\
\cmidrule{2-9}
& D- & B- & A- 
& PSNR$\uparrow$ & SSIM$\uparrow$ & LPIPS$\downarrow$ 
& PCC$\uparrow$ & $\bar{\psi}\downarrow$ \\
\midrule

Variant 1 & CNN & CNN & CNN & 22.31 & 88.69 & 12.48 & 0.58 & \textbf{12.44} \\
Variant 2 & Trans & Trans & CNN & 20.71 & 75.36 & 25.31 & \textbf{0.73} & 17.28 \\
Variant 3 & Trans & Trans & Trans & 20.57 & 75.00 & 25.73 & \underline{0.71} & 14.23 \\
Variant 4 & N/A & CNN & Trans & \underline{22.51} & \underline{88.85} & \underline{11.47} & \underline{0.71} & 15.95 \\

\midrule

Ours & CNN & CNN & Trans & \textbf{23.59} & \textbf{90.16} & \textbf{10.42} & 0.62 & \underline{12.83} \\

\bottomrule
\end{tabular}
}

\end{table}

\section{Conclusion}
\label{conclusion}
In this paper, we introduced a novel Physics-Aware Triple-Stream Underwater Image Enhancement Network, namely, PATS-UIENet, which predicts the degradation variables defined by the revised IFM, for the UIE task. To overcome the challenge of insufficient data, we also adopted an IFM-inspired semi-supervised learning framework, comprising a bi-directional supervised scheme and an unsupervised scheme. Due to the complementary action of the two schemes, the PATS-UIENet can be better trained using this framework with both the labeled and unlabeled real-world images while the generalization of the model is stronger, compared to supervised and unsupervised methods. To our knowledge, this study made the first effort to jointly exploit the physics-aware deep network and the IFM-inspired semi-supervised learning technique for the UIE task. The proposed method performed better than, or at least comparably to, sixteen baselines on six underwater testing sets in different evaluations. The promising results should be due to the fact that our method is able to not only estimate degradation parameters but also learn the characteristics of diverse underwater scenes.

In the future, we will further investigate more efficient architectures and training strategies to improve the performance of PATS-UIENet on high-resolution underwater images and facilitate its real-time deployment. In addition, we plan to incorporate more explicit physical priors and scene-adaptive degradation modeling to further enhance the interpretability and robustness of the proposed framework.

\section*{Acknowledgement}
This study was supported in part by the National Natural Science Foundation of China (No. 42576200) and the Key Research and Development Program of Shandong Province (No. 2024ZLGX06).

\section*{Data availability}
Data will be made available on request.

\section*{Declaration of competing interest}
The authors declare that they have no known competing financial interests or personal relationships that could have appeared to influence the work reported in this paper.

\section*{CRediT authorship contribution statement}
\textbf{Shixuan Xu:} Data curation, Formal analysis, Methodology, Software, Validation, Visualization, Writing - original draft;
\textbf{Hao Qi:} Data curation, Formal analysis, Methodology, Software, Validation, Writing - original draft;
\textbf{Wei Wang:} Methodology, Writing - review \& editing;
\textbf{Chao Huang:} Methodology, Writing – review \& editing;
\textbf{Jie Wen:} Methodology, Writing - review \& editing;
\textbf{Junyu Dong:} Funding acquisition, Methodology;
\textbf{Xinghui Dong:} Conceptualization, Funding acquisition, Investigation, Methodology, Project Administration, Resources, Supervision, Writing - Review \& Editing.

\bibliographystyle{IEEEtranN}
\bibliography{ref}

\end{document}